# Extraction of Sleep Information from Clinical Notes of Patients with Alzheimer's Disease Using Natural Language Processing

Sonish Sivarajkumar, MS[1], Thomas Yu Chow Tam, MS [2], Haneef Ahamed Mohammad, MS[2], Samuel Viggiano, MS[2], David Oniani, BS[2], Shyam Visweswaran, MD, PhD[1,3], Yanshan Wang, PhD[1,2,3*]

[1]Intelligent Systems Program, University of Pittsburgh, Pittsburgh, PA; [2]Department of Health Information Management, University of Pittsburgh, Pittsburgh, PA; [3]Department of Biomedical Informatics, University of Pittsburgh, Pittsburgh, PA

## Abstract

***Objective:*** *Alzheimer's Disease (AD) is the most common form of dementia in the United States. Sleep is one of the lifestyle-related factors that has been shown critical for optimal cognitive function in old age. However, there is a lack of research studying the association between sleep and AD incidence. A major bottleneck for conducting such research is that the traditional way to acquire sleep information is time-consuming, inefficient, non-scalable, and limited to patients' subjective experience.*

***Materials and Methods:*** *A gold standard dataset is created from manual annotation of 570 randomly sampled clinical note documents from the adSLEEP, a corpus of 192,000 de-identified clinical notes of 7,266 AD patients retrieved from the University of Pittsburgh Medical Center (UPMC). We developed a rule-based Natural Language Processing (NLP) algorithm, machine learning models, and Large Language Model(LLM)-based NLP algorithms to automate the extraction of sleep-related concepts, including*

[*] Corresponding author: Yanshan Wang, PhD, FAMIA; yanshan.wang@pitt.edu

*snoring, napping, sleep problem, bad sleep quality, daytime sleepiness, night wakings, and sleep duration, from the gold standard dataset*

***Results:*** *Rule-based NLP algorithm achieved the best performance of F1 across all sleep-related concepts. In terms of Positive Predictive Value (PPV), rule-based NLP algorithm achieved 1.00 for daytime sleepiness and sleep duration, machine learning models: 0.95 and for napping, 0.86 for bad sleep quality and 0.90 for snoring; and LLAMA2 with finetuning achieved PPV of 0.93 for Night Wakings, 0.89 for sleep problem, and 1.00 for sleep duration.*

***Discussion:*** *Although sleep information is infrequently documented in the clinical notes, the proposed rule-based NLP algorithm and LLM-based NLP algorithms still achieved promising results. In comparison, the machine learning-based approaches didn't achieve good results, which is due to the small size of sleep information in the training data.*

***Conclusion:*** *The results show that the rule-based NLP algorithm consistently achieved the best performance for all sleep concepts. This study focused on the clinical notes of patients with AD, but could be extended to general sleep information extraction for other diseases.*

## Introduction

Alzheimer's Disease (AD) is the most common form of dementia in the United States (U.S.), which affects at least 5.7 million Americans with a projected increase to 13.8 million by mid-century due to a global aging population[1-3]. In 2015, official death certificates recorded more than 110 thousand deaths from AD, making it the sixth leading cause of death in the U.S. and the fifth leading cause of death in Americans 65 years of age or older[1]. Unlike deaths from stroke and heart disease, which decreased between 2000 and 2015, deaths from AD increased 123%. By the year 2050, 13.8 million Americans are expected to have AD with an associated cost of $1.2 trillion U.S. dollars, not including unpaid caregiver hours. Postponing dementia onset by even one year could result in nine million fewer cases worldwide than predicted by 2050[4]

and a reduction in care costs. Therefore, early intervention to reduce the risk of AD will have a better population health impact.

Social and behavioral determinants of health (SDOH) are modifiable factors and offer opportunities for reducing the risk of AD[5]. Sleep is one of the lifestyle-related SDOH factors that has been shown critical for optimal cognitive function in old age[6]. However, the association between sleep and AD incidence is complex, as shown in the literature. On the one hand, previous studies suggest that sleep problems, such as insomnia[7], excessive daytime sleepiness[8, 9], snoring[10], sleep duration[11], poor sleep quality[12], and difficulties maintaining sleep[11], are associated with an increased risk of incident cognitive impairment and could be an early predictor of future AD dementia[13]. On the other hand, some studies find no association between sleep variables (e.g., sleep duration, sleep difficulties, and snoring) and cognitive function[14]. Moreover, a bi-directional relationship is seen between sleep and cognitive function decline in the elderly with underlying AD; in other words, AD also causes circadian and sleep disturbances. Despite a growing interest in studying sleep-AD relationship, longitudinal epidemiological research in a large cohort is still needed to understand the relationship. A major bottleneck for conducting such research is that the traditional way to acquire sleep and AD data through multi-year follow-ups is time-consuming, inefficient, non-scalable, and limited to patients' subjective experience.

Large volumes of electronic health records (EHRs) collected by healthcare organizations offer an opportunity to use a large sample size to investigate intervention outcomes in routine care, such as predictors of response, safety, comparative effectiveness, and health economic evaluations[15]. EHRs have become popular in AD research, such as resource use in AD care[16], comorbidities[17], case capture efficiency[18], and health disparities[19]. However, EHRs remain underused in collecting sleep information for AD research. A major barrier that hinders the use of sleep information from EHRs for AD research is that most of the sleep information in EHRs is embedded in clinical narratives. Natural language processing (NLP), a technique in computational linguistics that uses computational models for understanding natural language, has been used to extract meaningful information from clinical narratives[20]. However, there existed no NLP algorithms in the literature to extract sleep information from clinical notes, particularly for patients with AD, to the best of our knowledge. In this study, we developed different NLP algorithms: rule-based NLP, machine learning-based, and Large Language Model(LLM)-based NLP algorithms to automate the extraction of sleep-related concepts, including snoring, napping sleep problem, poor sleep quality, daytime sleepiness, night wakings, and sleep duration, from the clinical narratives of patients diagnosed with AD. We trained and validated the proposed models on the clinical notes retrieved from the University of Pittsburgh Medical Center (UPMC). The results show that the rule-based NLP algorithm achieved consistently the best performance for extracting all sleep concepts.

**Background**

Several prior studies have used the EHR data for the sleep information, most of which utilized structured EHR data, such as the International Classification of Diseases (ICD) diagnostic codes. For example, Felder et al. used ICD codes to identify sleep disorders and study the association between sleep disorders and preterm birth[21]. Hsiao et al.[22] identified patients with sleep disorders using ICD-9 codes 307.4 and 780.5x and explored the association between sleep disorders and autoimmune diseases. ICD codes have also been used to identify obstructive sleep apnea[23, 24]. However, it has been shown that sleep disorders are poorly coded in structured EHR data. ICD codes for identifying a sleep disorder from inpatient EHR data has only 79.2% sensitivity and 28.4% specificity[25]. A study also found that manual chart review of unstructured EHR data was able to identify 50% more individuals with insomnia and 68% more individuals with sleep problems, compared to using only ICD code[26]. Therefore, unstructured EHR data (e.g., clinical notes) is valuable for identifying sleep information.

Despite the rich sleep information embedded in unstructured EHRs, only a limited number of studies are found in the literature applying NLP and machine learning methods to automatically extract sleep information from unstructured EHRs. Divita et al.[27] applied a keyword matching approach to extract general symptoms that includes sleepiness from VA clinical notes. Similarly, a few studies used NLP to extract disturbed sleep and insomnia from clinical notes to study the association between sleep and mental disorders[28]. For example, Zhou et al.[29] used sleep-related symptoms to identify patients with depression, and Irving et al.[30] considered insomnia and distributed sleep in clinical notes to predict psychosis. Kartoun et al.[31] used text mention of sleep disorders in clinical notes to predict insomnia and found superior performance in identifying insomnia patients compared to ICD codes. Other studies used sleep-related reactions, such as sleeplessness and sleepy, from clinical notes or adverse event reports of adverse drug events[32, 33]. However, none of the studies focused on automated methods of extracting a comprehensive list of sleep variables related to the AD. Therefore, this study aims to use NLP and machine learning to automate the extraction of a comprehensive list of sleep-related variables.

## Materials and Methods

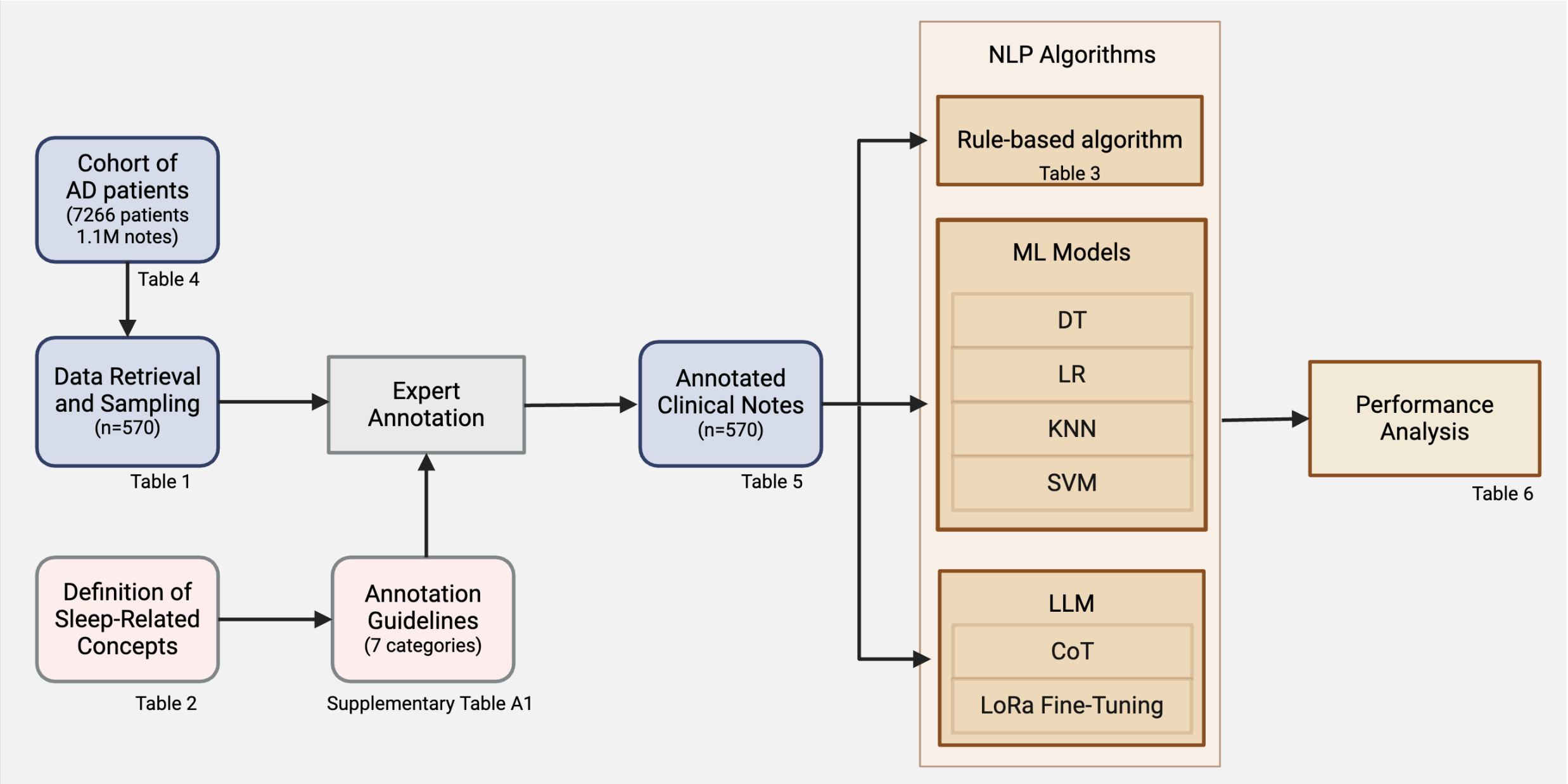

**Figure 1**: Workflow of the study

In Figure 1, we present the comprehensive workflow of our study, which outlines the systematic process from the initial identification of the AD patient cohort to the development and performance analysis of various NLP models for extracting sleep-related information from clinical notes.

*Data Collection*

We first defined a cohort of patients diagnosed with AD (ICD-10 codes: G30.0, G30.1, G30.8, and G30.9) between January 1st, 2020 and December 31st, 2020 at UPMC. We collected all their clinical notes that were created between January 1st, 2016 and December 31st, 2020, through the data service provided by the University of Pittsburgh Health Record Research Request (R3). The University of Pittsburgh's Institutional Review Board (IRB) reviewed and approved this study's protocol.

*Data Preprocessing*

A total of 7,266 patients were identified, and approximately 1.1 million de-identified clinical documents were retrieved. Since these clinical documents were auto-populated in the EHR system, we developed a data preprocessing algorithm to clean the clinical text. Specifically, we integrated clinical documents with the same document ID regardless of the line number ID and removed duplicated clinical documents. Although some documents are not duplicated, most contents in those documents are overlapped due to the nature of data population. For example, a physician digitally signing a clinical document will generate a duplicated document with just the addition of "Digitally signed by…" one or two days later. Therefore, we applied a surface lexical similarity approach to identify the highly similar documents. Suppose that V is a set of unique words that occurred in documents $D_1$ and $D_2$. $D_1$ and $D_2$ can be represented in the same vector space as $d_1$ and $d_2$ respectively where each component corresponds to the word in V and the value is the word frequency. Then we calculated the cosine similarity between two document vectors. If the similarity score is greater than 0.9, we will randomly remove one of the duplicated documents. After the data preprocessing, the total number of clinical documents was 379k.

Another challenge in extracting sleep information from clinical notes is that not every clinical note records relevant information. To identify relevant documents, we used information retrieval (IR) to select the documents. We applied fuzzy search that returns the documents containing keywords related to sleep regardless of the morphological format. The keywords used in our search are shown in the Appendix Table S1. As a result, 192k (51%) out of 379k documents were returned for further investigation in this project. This set of 192k documents is called adSLEEP corpus hereafter.

*Gold Standard Data Annotation*

We randomly sampled 570 clinical note documents from the adSLEEP corpus for manual annotation to create the gold standard dataset. Two health informatics students annotated this sampled dataset. The annotator was directed to annotate mentions of seven sleep-related categories in each clinical note, including snoring, napping, sleep problem, sleep quality, daytime sleepiness, night wakings, sleep duration. These sleep-related concept categories are defined based on previous sleep studies for patients with AD[6, 12, 34]. Table 1 shows the detailed definitions for each of the classes. Then a judge aggregated the concept annotations in a clinical note to a document-level label. Initially, a batch of 20 documents was given to the annotators to refine the annotation guidelines and to discuss the discrepancies to reach consensus on the concept definition. We repeated this process on another batch of 20 documents until an inter-annotator agreement (IAA) above 0.60 in terms of Cohen's Kappa was achieved. These 40 documents were used to measure a final IAA. Then we annotated the remaining 530 clinical notes using the updated annotation guidelines. Table S3 in the Appendix lists the detailed concept definitions in the annotation guidelines including tag names, keywords/phrases, and attributes.

**Table 1**. Definitions and examples of sleep-related concepts.

| **Concept Category** | **Definition** | **Example** |
|---|---|---|
| Snoring<br>(Yes or No) | Snoring or snoring synonyms | "snoring"<br>"snored" |
| Napping<br>(Yes or No) | Napping during daytime | "napping"<br>"doze" |
| Sleep problem<br>(Yes or No) | Sleep problem<br>Specific sleep disorder/ condition/disease mentioned in the note | "sleep disorder"<br>"insomnia"<br>"hypersomnia" |
| Bad sleep quality<br>(Yes or No) | Any mention related to bad sleep quality in the note | "sleeplessness"<br>"couldn't sleep during night"<br>"staying up all night" |
| Daytime sleepiness<br>(Yes or No) | Sleepiness during daytime | "sleep a lot through out the day"<br>"excessive daytime sleepiness" |
| Night wakings | Night time wakings | "frequent night waking" |
| (Yes or No) | | "waking up in the middle"<br>"waking up 3-5 times" |
| Sleep duration<br>(Short <=6h, Medium 6-8h, or Long >=8h) | Duration of night time sleep | "sleeps 4-5 hours" --> Short<br>"sleep more than 12 hours" --> Long |

*Rule-based NLP Algorithm*

We developed a rule-based NLP algorithm named *nlp4sleep* for sleep information extraction using MedTagger[35], a clinical NLP tool based on the Unstructured Information Management Architecture (UIMA) framework[36]. The MedTagger software is publicly available at GitHub[2].

We first used top-down and bottom-up approaches to identify the keywords in the rules for each sleep concept extraction. For the top-down approach, we searched the synonyms for each concept in the medical terminologies and ontologies, including Unified Medical Language System (UMLS) Metathesaurus. For the bottom-up approach, we used word embeddings trained from the clinical corpus to find the top 3 most similar terms. Then we used 70% (200 documents) of the gold standard dataset as training data to develop regular expression rules for the NLP algorithm. Since MedTagger includes the negation detection and hypothetical mention detection, we didn't specify negation rules unless we saw undetected negations in the training data. Table S2 in the Appendix lists the regular expression rules used in the *nlp4sleep* algorithm to extract sleep concepts. The NLP system extracted sleep concepts from each clinical document and assigned a document-level classification for each concept. If there were multiple mentions one of a concept in a document, we applied majority voting strategy to obtain the final document label. The  NLP algorithm is publicly available through the Open Health Natural Language Processing (OHNLP) consortium at GitHub (https://github.com/OHNLP/nlp4sleep).

[2] https://github.com/OHNLP/MedTagger

*Machine Learning Models*

In addition to the rule-based NLP algorithm, we also developed machine learning models to extract sleep-related concepts. Four major machine learning-based clinical text classification models, namely encompassing Decision Trees(DT), Logistic Regression (LR), K-Nearest Neighbors (KNN), and Support Vector Machine (SVM) were trained and tested for each sleep concept. Since real-world clinical text data may be inadequate and inconsistent for machine learning-based NLP models to comprehend, we adopted multiple preprocessing steps before feeding the text data to the machine learning models. After converting into lower case, the text was tokenized to break the sentences into tokens, including words, phrases, symbols, or other meaningful elements. Stop words and non-numeric were removed from the token lists. The tokens were then lemmatized to the base forms to reduce the complexity of the text. Then we converted the entire document to a numeric vector using Term Frequency-Inverse Document Frequency (TF-IDF) vectorization. All experiments, with the exception of 'Sleep Duration', were performed as a binary text classification task, assigning positive or negative predictions to each concept category. 'Sleep Duration', however, was treated as a multi-label classification task.. The same training and test datasets were used as in the rule-based NLP algorithm, and the performance on the test dataset was reported.

*LLM-based algorithms*

We further extended our computational methodologies for sleep information extraction by incorporating LLMs with a focus on LLAMA2[37], leveraging both Chain-of-Thought(CoT) prompting and Supervised Fine-Tuning(SFT) approaches. The decision to utilize LLAMA2 stems from its demonstrated effectiveness in clinical information extraction tasks, as evidenced in our prior research works[38, 39], and its availability as an open-source model aligns with our commitment to accessible and reproducible scientific work.

The Chain of Thought (CoT) based NLP algorithm was implemented through carefully crafted prompts designed for each sleep concept. This approach harnesses the model's capability to generate intermediate reasoning steps, facilitating a more nuanced understanding of complex clinical narratives found in EHRs. By simulating a thought process that mirrors clinical reasoning, CoT prompting aims to improve the model's accuracy in identifying and classifying the various sleep concepts. The customized prompts were developed based on insights from clinical experts and iteratively refined to capture the intricacies of sleep-related terminology and contexts within the clinical notes.

Beyond the CoT prompting, we also implemented Supervised Fine-Tuning (SFT) on LLAMA2, specifically adopting LoRA-based tuning techniques known for their efficiency in parameter adjustment. This approach fine-tunes a select group of parameters within LLAMA2, resulting in a model variant precisely calibrated for identifying sleep-related concepts within EHR data. The SFT methodology involved instruction tuning the model on a labeled dataset comprising examples of each sleep concept, enabling the model to learn from real-world clinical narratives and apply these learnings to accurately classify new, unseen data.

*Evaluation*

We used 30% (120 documents) of the gold standard dataset as the testing data to validate the rule-based NLP algorithm and machine learning models. All models were evaluated using sensitivity, specificity, positive predictive value (PPV), and F1 score. Since our dataset is imbalanced, we report the weighted-averaged F1 score. The definitions of the evaluation metrics are shown below:

$$Sensitivity = \frac{True\ Positive}{True\ Positive + False\ Nagative}$$

$$Specificity = \frac{True\ Nagative}{True\ Nagative + False\ Positive}$$

$$PPV = \frac{True\ Positive}{True\ Positive + False\ Positive}$$

$$F1\ score = \frac{2 \cdot True\ Positive}{2 \cdot True\ Positive + False\ Positive + False\ Nagative}$$

**Results**

**Table 2** shows the demographics of the AD cohort of 7,266 patients. Patients were primarily white (91%), female (64%), and not Hispanic or Latino (95%), with a mean age of 85 years. The demographics of this cohort are similar to the demographics of the population with AD in western Pennsylvania.

**Table 3** lists the number of documents for each sleep concept in the annotated training and test datasets. As shown in the table, the frequency of these sleep concepts is low in the gold standard dataset. Though the clinical documents were identified by using a list of relevant keywords, most documents do not contain any sleep-related concepts. The reason might be that some keywords may not be only related to sleep; for example, wheezing might be related to respiratory diseases.

**Table 2**. Demographics of the AD

| Demographics | Total (n=7,266) |
|---|---|
| Age (Mean) | 85 |
| **Sex** | |
| Female | 4,649 (64%) |
| Male | 2,617 (36%) |
| **Race** | |
| White | 6,628 (91%) |
| Black | 482 (6.6%) |
| Asian | 67 (1.1%) |
| Others | 67 (1.1%) |
| Not Specified | 22 (0.3%) |
| **Ethnicity** | |
| Hispanic or Latino | 33 (0.5%) |
| Not Hispanic or Latino | 6,900 (95%) |
| Not Specified | 333 (4.5%) |

The performance of the rule-based NLP algorithm and machine learning models is listed in **Table 4**. The rule-based NLP algorithm demonstrated exceptional performance across all sleep-related concepts, achieving perfect scores in sensitivity, specificity, F1, and PPV for the concept of daytime sleepiness and sleep duration. This algorithm's strength lies in its ability to accurately identify instances of sleep concepts with precision, as evidenced by its high PPV values, notably in the snoring concept (0.94) and sleep duration where it also achieved perfect scores. However, its performance varied across other sleep concepts, with the lowest scores observed in napping (specificity and PPV both at 0.5), indicating potential challenges in distinguishing relevant napping instances from unrelated contexts. Despite these variances, the rule-based approach excelled in extracting specific sleep-related concepts, particularly in accurately identifying instances without false negatives, as shown by the perfect sensitivity scores across the board.

The machine learning models, encompassing Decision Trees (DT), Logistic Regression (LR), K-Nearest Neighbors (KNN), and Support Vector Machine (SVM), showed varied performances across different sleep concepts. The SVM model, in particular, demonstrated robustness, with high sensitivity and specificity scores, peaking at 0.95 for identifying night wakings and maintaining strong performance in sleep duration. However, these models generally exhibited lower PPV scores compared to the rule-based NLP algorithm, indicating a higher rate of false positives. The KNN model displayed notable consistency across metrics, suggesting its capability in handling the unstructured nature of clinical text data, albeit with some limitations in precision as indicated by its PPV scores. The variability in performance across these models underscores the challenges in applying machine learning to clinical text classification, particularly with sparse and infrequent concepts. These results are consistent with previous studies[40, 41] that machine learning models might not be effective in clinical text classification when the size of the annotated training dataset is small, and the concepts of interest are sparse and infrequent in the documents.

**Table 3**. Number of clinical documents for each sleep concept in the annotated training

| **Concept Category** | **# Documents in Training (Yes/No)** | **# Documents in Test (Yes/No)** |
|---|---|---|
| Snoring | 60 / 290 | 21 / 199 |
| Napping | 31 / 319 | 10 / 210 |
| Sleep problem | 71 / 279 | 23 / 197 |
| Bad sleep quality | 45 / 305 | 25 / 195 |
| Daytime sleepiness | 103 / 247 | 34/ 186 |
| Night wakings | 104 / 246 | 35 / 115 |
| Sleep duration | 240(Short) / 47(Medium)/ 67(Long) | 80(Short) / 13(Medium) / 22(Long) |

**Table 4**. Performance of the rule-based NLP algorithm and machine learning models.

| **Sensitivity<br>Specificity<br>F1<br>PPV** | **Daytime Sleepiness** | **Napping** | **Night Wakings** | **Sleep Problem** | **Bad Sleep Quality** | **Snoring** | **Sleep Duration** |
|---|---|---|---|---|---|---|---|
| Rule-based NLP | **1.00**<br>**1.00**<br>**1.00**<br>**1.00** | 0.5<br>**0.99**<br>**0.98**<br>0.5 | **1.00**<br>**0.99**<br>**0.99**<br>0.75 | 0.85<br>**0.93**<br>**0.91**<br>0.8 | 0.62<br>0.51<br>**0.91**<br>0.6 | 0.94<br>**0.97**<br>**0.97**<br>0.89 | **1.00**<br>**1.00**<br>**1.00**<br>**1.00** |
| DT | 0.86<br>0.86<br>0.90<br>0.86 | **0.89**<br>0.89<br>0.98<br>0.89 | 0.82<br>0.80<br>0.81<br>0.84 | 0.78<br>0.72<br>0.74<br>0.84 | 0.63<br>0.57<br>0.58<br>0.81 | 0.75<br>0.75<br>0.75<br>0.78 | 0.79<br>0.74<br>0.76<br>0.77 |
| LR | 0.90<br>0.77<br>0.81<br>0.89 | 0.47<br>0.50<br>0.48<br>0.94 | 0.92<br>0.83<br>0.86<br>0.91 | **0.91**<br>0.58<br>0.59<br>0.82 | 0.42<br>0.50<br>0.45<br>0.83 | 0.42<br>0.50<br>0.46<br>0.84 | 0.77<br>0.67<br>0.71<br>0.78 |
| KNN | 0.93<br>0.85<br>0.88<br>0.93 | 0.79<br>0.79<br>0.79<br>**0.95** | 0.87<br>0.79<br>0.82<br>0.89 | 0.79<br>0.65<br>0.68<br>0.83 | 0.76<br>0.72<br>0.74<br>**0.86** | 0.70<br>0.66<br>0.71<br>0.81 | 0.84<br>0.89<br>0.86<br>0.87 |
| SVM | 0.91<br>0.80<br>0.84<br>0.90 | 0.81<br>0.69<br>0.74<br>**0.95** | 0.90<br>0.85<br>0.87<br>0.91 | 0.91<br>0.61<br>0.63<br>0.83 | **0.93**<br>0.60<br>0.63<br>**0.86** | **0.95**<br>0.69<br>0.75<br>**0.90** | 0.85<br>0.84<br>0.84<br>0.78 |
| LLAMA2-CoT | 0.69<br>0.54 | 0.72<br>0.62 | 0.48<br>0.34 | 0.82<br>0.86 | 0.74<br>0.67 | 0.8<br>0.85 | 0.77<br>0.72 |
| | 0.57<br>0.543 | 0.58<br>0.623 | 0.39<br>0.34 | 0.83<br>0.86 | 0.7<br>0.672 | 0.79<br>0.846 | 0.67<br>0.719 |
| LLAMA2-SFT | 0.93<br>0.92<br>0.91<br>0.92 | 0.82<br>0.94<br>0.88<br>0.85 | 0.9<br>0.946<br>0.96<br>**0.93** | 0.9<br>0.904<br>0.89<br>**0.89** | 0.87<br>0.871<br>0.84<br>0.84 | 0.88<br>0.88<br>0.78<br>0.83 | **1.00**<br>**1.00**<br>**1.00**<br>**1.00** |

The LLM-based NLP algorithms, LLAMA2-Chain of Thought (CoT) and LLAMA2 with finetuning (SFT), introduced advanced contextual understanding to the task. The LLAMA2-SFT, leveraging LoRA-based parameter-efficient finetuning, exhibited remarkable performance, closely rivaling the rule-based NLP algorithm, particularly in sleep duration where it achieved perfect scores. It achieved high sensitivity, specificity, and F1 scores, especially in processing complex sleep concepts like night wakings and sleep problems, indicating its strong contextual comprehension and adaptability. The LLAMA2-CoT approach showed a more moderate performance, illustrating the potential limitations of relying solely on reasoning chains without finetuning for highly specialized tasks like clinical concept extraction.

Comparing the three types of algorithms, the rule-based NLP and LLAMA2-SFT models stand out for their superior performance, particularly in capturing the intricacies of sleep duration with perfect accuracy. The rule-based NLP algorithm excels in specificity and sensitivity, attributed to its tailored rules that effectively capture specific sleep-related concepts. LLAMA2-SFT, with its finetuning, demonstrates comparable excellence, benefiting from the deep contextual understanding and adaptability of LLMs to the nuances of clinical narratives. While machine learning models offer valuable insights, their performance, particularly in terms of PPV, indicates a susceptibility to false positives, a critical limitation in clinical applications where accuracy is paramount.

The exceptional results of the rule-based NLP and LLAMA2-SFT models underscore their effectiveness in clinical text classification, suggesting that a hybrid approach leveraging the precision of rule-based methods and the contextual adaptability of finetuned LLMs could provide a robust solution for extracting sleep information from the unstructured text of EHRs. This is particularly evident in their handling of sleep duration, where both models demonstrated their capability to accurately and reliably capture sleep patterns, highlighting the potential for comprehensive sleep information extraction in clinical settings.

*Error Analysis of the rule-based NLP algorithm*

We conducted an error analysis of the documents misclassified by the rule-based NLP algorithm and analyzed the causes of the false positives and false negatives for each sleep concept. Some false positives were due to our annotators failing to annotate the information. For example, in the text "Histories Past Medical History Combined Chronic Systolic/Diastolic CHF COPD (industrial exposure) CAD s/p stents PMH Right Ocular Stroke - chronic visual defect BPH Type 2 DM OSA on BiPAP", the rule-based NLP algorithm identified OSA as snoring and sleep problem. However, the concept had not been annotated. Many semi-structured clinical text auto-populated in the EHR system is difficult for the annotator to read and annotate. In another false positive, the sentence "He had episodes yesterday in which he became confused after waking up from a nap" indicates that the patient had a nap that may not be related to sleep pattern. In another false positive case for sleep problem "Depression screen done 7/2017, PHQ9 score 16 points for sleep problem which seems better now", the NLP algorithm couldn't identify that this sentence was not about a positive sleep problem. In a false positive case for bad sleep quality, the sentence "Take Melatonin 5 mg at bedtime every night for 3- 4 weeks for difficulty falling asleep" was a suggestion for the patient.

Some false negatives are due to errors in negation detection. For example, in the sentence "The patient's daughter states that she has not been complaining of her back pain or of her leg cramps we discussed the fact that she is doing less and does nap during the day", the algorithm incorrectly identified this mention as negated since it failed to identify two sentences. In another example for sleep problem "Change in social contacts/activities? No Patient Active Problem List Diagnosis Primary open angle glaucoma Urge incontinence Backache, unspecified Pneumonia, organism unspecified Insomnia", the NLP algorithm incorrectly split the sentence into "Change in social contacts/activities?" and "No Patient Active Problem

List Diagnosis Primary open angle glaucoma Urge incontinence Backache, unspecified Pneumonia, organism unspecified Insomnia" and wrongly identified the negation. The semi-structured clinical text also confused the NLP algorithm in detecting sentences and negations.

**Discussion**

Detailed descriptions of SDOH are usually captured in unstructured clinical text; however, the SDOH information may be sparsely documented due to the lack of clinical practice guidelines for documenting such information. Our study shows that sleep information is infrequently recorded in clinical notes for patients with AD. For example, in the gold standard dataset, only 14% of clinical documents recorded snoring concepts, 1.6% napping, 17.2% sleep problem, 8.8% bad sleep quality, 3.1% daytime sleepiness, 1.3% night wakings, and 1% sleep duration. This observation raises the question of whether such under-documented SDOH information in clinical notes will be useful for downstream statistical analysis in a cohort study or an epidemiology study.

Another challenge we encountered during the project was the definition of sleep-related concepts. We initially considered eight concepts, including sleep disorder, sleep problem symptoms, snoring, napping, sleep quality, daytime sleepiness, night wakings, and sleep duration, with detailed granularity according to the relevant sleep research in the literature. For example, there were four categories associated with snoring or daytime sleepiness: negated, positive, sometimes, and all the time. There were three categories for night wakings: 0, 1-2, and >2. However, during the annotation process, we found that the granular categories for each concept were rarely used and there were significant overlaps between sleep problem symptoms and other concepts. For example, phrases like "staying up all night" meet the description of insomnia, but the patient was never diagnosed with insomnia. Likewise, snoring and sleep apnea shared concept-likeness but are not always annotated similarly. For example, the concept of snoring is annotated as snoring but not sleep apnea. However, the concept of sleep apnea is annotated as a sleep problem and snoring because the concept meets both definitions. Thus, we simplified the concept definition and the categories for each concept.

Since SDOH comprises more conditions related to socioeconomic status, living environment, housing, education, food, community, it might be more challenging to define these SDOH concepts. It is also questionable whether such information is adequately documented in EHRs and whether such information from EHRs would be useful for research. Thus, a feasibility study of assessing the availability of SDOH in EHRs for a certain cohort of patients might be necessary before algorithm development. In addition, it is also a tedious and time-consuming process to manually annotate a gold standard dataset. A potential beginning point of building automated systems to extract SDOH from EHRs might be a community effort to build an SDOH ontology and terminology.

Additionally, sleep information is infrequently documented in the clinical notes and keywords are shared with concepts. Although we used IR to select the documents with keywords related to sleep, keywords including wheeze, wheezing, and apnea appear often but are unrelated to the patient's sleep. For example, physicians commonly check a patient's respiratory health and record the presence of wheezing in the clinical notes. Wheezing, a shared concept between respiratory health and sleep, was found problematic when retrieving sleep-specific documents. Using a comprehensive list of keywords that sufficiently cover the domain of interest to retrieve relevant clinical documents has been adopted as the general approach for sampling a set of documents to be annotated and extracted. However, this approach requires tedious work with collaboration with an engaged focus group. In addition, this sampling approach hampers the NLP methods to identify rare cases and uncommon phenotypes, which is a major threat to NLP generalizability.

*Limitations*

There are several limitations in this study. First, the ICD codes used to define AD may not be optimal. However, a more comprehensive way to define AD is out of scope of this study. Second, the initial search keywords used to retrieve sleep-related clinical notes may not be complete and could miss some documents. However, this could be a common problem for SDOH information extraction due to the sparse and infrequent documentation in clinical notes. Third, we acknowledge that the annotated dataset used to train and test the proposed systems is relatively small, which may limit the usability of the system and discredit the conclusions. However, the clinical note annotation is a time-consuming and expensive process. Each document requires substantial time (~2 hours) for each annotator to complete the annotations of seven sleep-related concepts. The proposed NLP systems in this study are still valuable to the literature. Last but not least, we didn't consider sleep information in other EHR data types (e.g., diagnosis codes, survey data, questionnaire data), sleep studies such as polysomnography, and sleep tests such as multiple sleep latency test (MSLT), which should be considered for further cohort studies on the association between sleep and AD.

*Future Work*

In future work, we plan to explore more sophisticated methods in retrieving relevant documents to the considered medical concepts with high precision. This might be a key challenge in collecting a corpus for studying an SDOH concept. In addition, we will also investigate novel machine learning methods that require less or no data for training, such as semi-supervised learning and self-supervised learning.

This work could also nicely support the research on the connection between sleep and AD. Knowing that sleep is one of the modifiable lifestyle-related factors, this provides evidence that the research being conducted on AD and sleep interventions is necessary and critical. Research should continue to understand the associations among sleep variables (e.g., sleep duration, sleep difficulties, and snoring) and cognitive function, as well as interventions that are more effective to address sleep disturbances in older adults with AD.

**Conclusion**

The study underscores the effectiveness of NLP in extracting sleep information from the clinical notes of AD patients, with the rule-based algorithm showing the highest accuracy across all sleep concepts. Our findings demonstrate that the rule-based NLP algorithm consistently outperformed machine learning and LLM-based algorithms across all evaluated sleep concepts, showcasing its superior accuracy and reliability.

This study focused on the clinical notes of patients with AD, but could be extended to general sleep information extraction for other diseases.

Furthermore, the methodologies and findings of this study have broader implications for the application of NLP in healthcare. The open-source nature of the developed rule-based NLP algorithm and the insights gained from comparing different NLP approaches can be leveraged by other researchers and practitioners to advance the extraction of health-related information from EHRs.

**Acknowledgements**

This project was partially supported by the University of Pittsburgh Momentum Funds and the National Institutes of Health through Grant Number UL1TR001857 and U24TR004111. The funders had no role in the design of the study, and collection, analysis, and interpretation of data and in preparation of the manuscript. The views presented in this report are not necessarily representative of the funder's views and belong solely to the authors.

## APPENDIX

**Table S1**. Sleep-related keywords used to retrieve relevant clinical note documents.

| |
|---|
| 'snore', 'snoring', 'wheeze', 'wheezing', 'sleep', 'sleepiness', 'sleeping', 'sleepless', 'sleeplessness', 'apnea', 'hypopnea', 'osa', 'insomnia', 'nap', 'napping', 'narcolepsy', 'nocturnal', 'somnolence', 'somnolent', 'dizziness', 'hypersomnia', 'rem', 'nrem', 'wake', 'wakefulness', 'waking', 'polysomnography' |

**Table S2**. Regular expression rules used in the NLP algorithm for the extraction of sleep concepts from clinical notes.

<table>
<tr><th>Concept Category</th><th>Keywords and Regular Expressions</th></tr>
<tr><td>Snoring</td><td>snor(es|ing|e)?; snorings; sleep apnea; osa; obstructive sleep apnea</td></tr>
<tr><td>Napping</td><td>nap(s|ping)?</td></tr>
<tr><td>Sleep problem</td><td>insomnia; sleeplessness; sleep (disorders?|problems?); hypersomnia; parasomnia; osa; obstructive sleep apnea; sleep apnea; hypersomnolence</td></tr>
<tr><td>Bad sleep quality</td><td>staying up; (trouble|irritable|tense) (\S+\s+){0,5}(sleep(ing)?|asleep); sleep(s|ing)? poorly; sleep is poor; restless sleep; ca(n't|nnot) sleep; sleep issues?; sleep(ing)? (\S+\s+){0,5}(problems?|problematic); sleeps? a lot; difficulty (\S+\s+){0,5}(asleep|sleep(ing)?); sleep disturbance; disturbance in sleep; sleep quality: (fair|bad); not sleeping; no sleep; sleep difficulty; nocturnal agitation; up (during|at) night; nocturnal; often awake</td></tr>
</table>

<table>
<tr><td>Daytime sleepiness</td><td>(excessive )?daytime sleep(iness|inesses)?; (excessive )?daytime somnolence; sleep(s|ing|iness)? at times; sleep(s|iness)? in (\S+\s+){0,2}day(time)?; sleep(s|iness)? during (\S+\s+){0,2}day(time)?; sleep all day</td></tr>
<tr><td>Night wakings</td><td>night (wakings|awakenings), wak(e|es|ing up) (\S+\s+){0,5}night; awake(ning|n)? (from|during|at) night(mares)?</td></tr>
<tr><td>Sleep duration</td><td>Short: sleep(s|ing)? (less than|up to) (1|2|3|4|5|6) hours<br><br>Medium: sleep(s|ing)? (\S+\s+){0,5}(6|7|8)-(6|7|8) hours<br><br>Long: sleep(s|ing)? (\S+\s+){0,5}more than (8|9|10|11|12|13|14|15|16|17|18|19|20) hours; sleep(s|ing)? (\S+\s+){0,5}(8|9|10|11|12)-(8|10|11|12|13|14|15|16|17|18|19|20) hours</td></tr>
</table>

**Table S3**. Concept Definitions in the Annotation Guidelines including Tag Names, Keywords/Phrases, and Attributes

| Tag Name/Concepts | Example Keywords/Phrases | Attributes | |
|---|---|---|---|
| Snoring<br>*Snoring or snoring synonyms* | "snoring"<br>"wheezing"<br>"sleep apnea" | *type* | negated<br>positive |
| | | | |
| Napping<br>*Napping during daytime* | "napping"<br>"doze" | *type* | negated<br>positive |
| | | | |
| Sleep problem<br>*Sleep disorder/ condition/disease mentioned in the note* | “sleep disorder”,<br>“insomnia”,<br>“hypersomnia”,<br>“couldn’t sleep during night”<br>“staying up on night” | *type* | negated<br>positive |
| | | | |
| Sleep quality<br>*Sleep quality mentioned in the note* | Good – sleep quality is good, satisfaction with the sleep experience | *type* | good<br>bad |
| | | | |
| Daytime Sleepiness<br>*Sleepiness during daytime* | “sleep a lot through out the day”<br>“excessive daytime sleepiness” | *type* | negated<br>positive |
| | | | |
| Night wakings<br>*Night time wakings* | “frequent night waking”<br>“waking up in the middle”<br>“waking up 3-5 times” | *type* | Negated<br>Positive |
| | | | |
| Sleep duration<br>*Duration of night time sleep* | “sleeps 4-5 hours”<br>“more than 2 hours at a time is reported as good” | *type* | Short (<6.5h)<br>Medium (6.5-8.5h)<br>Long (>8.5h) |